# Identifying Hierarchical Structure in Sequences:
# A linear-time algorithm

**Craig G. Nevill-Manning**                                          CGN@CS.WAIKATO.AC.NZ
**Ian H. Witten**                                                    IHW@CS.WAIKATO.AC.NZ
*Department of Computer Science*
*University of Waikato, Hamilton, New Zealand.*

**Abstract**

SEQUITUR is an algorithm that infers a hierarchical structure from a sequence of discrete symbols by replacing repeated phrases with a grammatical rule that generates the phrase, and continuing this process recursively. The result is a hierarchical representation of the original sequence, which offers insights into its lexical structure. The algorithm is driven by two constraints that reduce the size of the grammar, and produce structure as a by-product. SEQUITUR breaks new ground by operating incrementally. Moreover, the method's simple structure permits a proof that it operates in space and time that is linear in the size of the input. Our implementation can process 50,000 symbols per second and has been applied to an extensive range of real world sequences.

## 1. Introduction

Many sequences of discrete symbols exhibit natural hierarchical structure. Text is made up of paragraphs, sentences, phrases, and words. Music is composed from major sections, motifs, bars, and notes. Records of user interface behavior encode the hierarchical structure of tasks that users perform. Computer programs constitute modules, procedures, and statements. Discovering the natural structure that underlies sequences is a challenging and interesting problem that has a wide range of applications, from phrase discovery to music analysis, from programming by demonstration to code optimization.

    The search for structure in sequences occurs in many different fields. Adaptive text compression seeks models of sequences that can be used to predict upcoming symbols so that they can be encoded efficiently (Bell *et al.*, 1990). However, text compression models are extremely opaque, and do not illuminate any hierarchical structure in the sequence. Grammatical inference techniques induce grammars from a set of example sentences, possibly along with a set of negative examples (Gold, 1967; Angluin, 1982; Berwick and Pilato, 1987). However, it is crucial to their operation that the input is not a continuous stream but is segmented into sentences, which are, in effect, independent examples of the structure being sought. A brief review of pertinent systems appears in Section 8. Techniques of Markov modeling and hidden Markov modeling make no attempt to abstract information in hierarchical form (Rabiner and Juang, 1986, Laird and Saul, 1994). Sequence learning also occurs in areas such as automaton modeling (Gaines, 1976), adaptive systems (Andreae, 1977), programming by demonstration (Cypher, 1993), and human performance studies (Cohen *et al.*, 1990), but generally plays only a peripheral role.

    In this paper we describe SEQUITUR, an algorithm that infers a hierarchical structure from a sequence of discrete symbols. The basic insight is that phrases which appear more than once can be replaced by a grammatical rule that generates the phrase, and that this process can be continued recursively, producing a hierarchical representation of the original sequence. The result

is not strictly a grammar, for the rules are not generalized and generate only one string. (It does provide a good basis for inferring a grammar, but that is beyond the scope of this paper.) A scheme that resembles the one developed here arose from work in language acquisition (Wolff, 1975, 1977, 1980, 1982), but it operated in time that is quadratic with respect to the length of the input sequence, whereas the algorithm we describe takes linear time. This has let us investigate sequences containing several million tokens—in previous work the examples were much smaller, the largest mentioned being a few thousand tokens. Another difference, which is of crucial importance in some practical applications, is that the new algorithm works incrementally. We return to Wolff's scheme, and compare it with SEQUITUR, in Section 8.

The ability to deal easily with long sequences has greatly extended the range of SEQUITUR's application. We have applied it to artificially-generated fractal-like sequences produced by L-systems, and, along with a unification-based rule generalizer, used it to recover the original L-system. The same method has inferred relatively compact deterministic, context-free grammars for million-symbol sequences representing biological objects obtained from stochastic, context-sensitive, L-systems, which has in turn greatly speeded the graphical rendering of such objects. We have applied SEQUITUR to 40 Mbyte segments of a digital library to generate hierarchical phrase indexes for the text, which provides a novel method of browsing (Nevill-Manning *et al.*, 1997). The algorithm compresses multi-megabyte DNA sequences more effectively than other general-purpose compression algorithms. Finally, with some post-processing, it has elicited structure from a two million word extract of a genealogical database, successfully identifying the structure of the database and compressing it much more efficiently than the best known algorithms. We touch on some of these applications in Section 3 below; Nevill-Manning (1996) describes them all.

This paper describes the SEQUITUR algorithm and evaluates its behavior. The next section gives a concise description of the algorithm in terms of constraints on the form of the output grammar. Section 3 gives a taste of the kind of hierarchies that SEQUITUR is capable of inferring from realistic sequences. Section 4 describes the implementation in more detail, with particular emphasis on how it achieves efficiency. Section 5 shows that the run time and storage requirements are linear in the number of input symbols, while Section 6 discusses the algorithm's behavior on extreme input strings. We end with a quantitative analysis of SEQUITUR's performance on several example sequences, and a review of related research.

## 2. The SEQUITUR Algorithm

SEQUITUR forms a grammar from a sequence based on repeated phrases in that sequence. Each repetition gives rise to a rule in the grammar, and the repeated subsequence is replaced by a non-terminal symbol, producing a more concise representation of the overall sequence. It is this pursuit of brevity that drives the algorithm to form and maintain the grammar, and, as a by-product, provide a hierarchical structure for the sequence.

At the left of Figure 1a is a sequence that contains the repeating string *bc*. Note that the sequence is already a grammar—a trivial one with a single rule. To compress it, SEQUITUR forms a new rule $A \rightarrow bc$, and *A* replaces both occurrences of *bc*. The new grammar appears at the right of Figure 1a.

The sequence in Figure 1b shows how rules can be reused in longer rules. The longer sequence consists of two copies of the sequence in Figure 1a. Since it represents an exact repetition, compression can be achieved by forming the rule $A \rightarrow abcdbc$ to replace both halves of the sequence. Further gains can be made by forming rule $B \rightarrow bc$ to compress rule $A$. This demonstrates the advantage of treating the sequence, rule $S$, as part of the grammar—rules may be formed from rule $A$ in an analogous way to rules formed from rule $S$. These rules within rules constitute the grammar's hierarchical structure.

The grammars in Figures 1a and 1b share two properties:
$p_1$: no pair of adjacent symbols appears more than once in the grammar;
$p_2$: every rule is used more than once.

Property $p_1$ requires that every digram in the grammar be unique, and will be referred to as *digram uniqueness*. Property $p_2$ ensures that each rule is useful, and will be called *rule utility*. These two constraints exactly characterize the grammars that SEQUITUR generates.

Figure 1c shows what happens when these properties are violated. The first grammar contains two occurrences of *bc*, so $p_1$ does not hold. This introduces redundancy because *bc* appears twice. In the second grammar, rule $B$ is used only once, so $p_2$ does not hold. If it were removed, the grammar would become more concise. The grammars in Figures 1a and 1b are the only ones for which both properties hold for each sequence. However, there is not always a unique grammar with these properties. For example, the sequence in Figure 1d can be represented by both of the grammars on its right, and they both obey $p_1$ and $p_2$. We deem either grammar to be acceptable. Repetitions cannot overlap, so the string *aaa* does not give rise to any rule, despite containing two digrams *aa*.

SEQUITUR's operation consists of ensuring that both properties hold. When describing the algorithm, the properties act as *constraints*. The algorithm operates by enforcing the constraints on a grammar: when the digram uniqueness constraint is violated, a new rule is formed, and when the rule utility constraint is violated, the useless rule is deleted. The next two subsections describe how this occurs.

**2.1 Digram Uniqueness**

When SEQUITUR observes a new symbol, it appends it to rule $S$. The last two symbols of rule $S$—the new symbol and its predecessor—form a new digram. If this digram occurs elsewhere in the grammar, the first constraint has been violated. To restore it, a new rule is formed with the digram on the right-hand side, headed by a new non-terminal symbol. The two original digrams are replaced by this non-terminal symbol.

|   | Sequence | Grammar |   | Sequence | Grammar |
|---|---|---|---|---|---|
| a | S → abcdbc | S → aAdA<br>A → bc | b | S → abcdbcabcdbc | S → AA<br>A → aBdB<br>B → bc |
| c | S → abcdbcabcdbc | S → AA<br>A → abcdbc | d | S → aabaaab | S → AaA<br>A → aab |
|   |   | S → CC<br>A → bc<br>B → aA<br>C → BdA |   |   | S → AbAab<br>A → aa |

**Figure 1** Example sequences and grammars that reproduce them: (a) a sequence with one repetition; (b) a sequence with a nested repetition; (c) two grammars that violate the two constraints; (d) two different grammars for the same sequence that obey the constraints.

NEVILL-MANNING & WITTEN

Table 1 shows the grammars that result when successive symbols of the sequence *abcdbcabcd* are processed. The second column shows the sequence observed so far, the third column gives the grammar created from the sequence, and the fourth column notes constraints that have been violated, and actions that are taken to resolve the violations.

When SEQUITUR adds the final *c* at symbol 6, the digram *bc* appears twice. SEQUITUR creates the new rule *A*, with *bc* as its right-hand side, and replaces the two occurrences of *bc* by *A*. This illustrates the basic procedure for dealing with duplicate digrams.

The appearance of a duplicate digram does not always result in a new rule. If the new digram appears as the right-hand side of an existing rule, then no new rule need be created: the non-terminal symbol that heads the existing rule replaces the digram. After symbol 9 in Table 1, a third *bc* appears, and the existing non-terminal symbol *A* replaces the third occurrence of *bc*. This results in a new pair of repeating digrams, *aA*, shown in the next line of Table 1. SEQUITUR accordingly forms a new rule *B*, which replaces the two occurrences of *aA*. SEQUITUR creates and maintains the hierarchy by an iterative process: the substitution of *A* for *bc* results in the new digram *aA*, which is itself replaced by *B*. For larger sequences, these changes ripple through the grammar, forming and matching longer rules higher in the hierarchy.

| symbol number | the string so far | resulting grammar | remarks |
|---|---|---|---|
| 1 | a | S → a | |
| 2 | ab | S → ab | |
| 3 | abc | S → abc | |
| 4 | abcd | S → abcd | |
| 5 | abcdb | S → abcdb | |
| 6 | abcdbc | S → abcdbc | bc appears twice |
| | | S → aAdA<br>A → bc | enforce digram uniqueness |
| 7 | abcdbca | S → aAdAa<br>A → bc | |
| 8 | abcdbcab | S → aAdAab<br>A → bc | |
| 9 | abcdbcabc | S → aAdAabc<br>A → bc | bc appears twice |
| | | S → aAdAaA<br>A → bc | enforce digram uniqueness.<br>aA appears twice |
| | | S → BdAB<br>A → bc<br>B → aA | enforce digram uniqueness |
| 10 | abcdbcabcd | S → BdABd<br>A → bc<br>B → aA | Bd appears twice |
| | | S → CAC<br>A → bc<br>B → aA<br>C → Bd | enforce digram uniqueness.<br>B is only used once |
| | | S → CAC<br>A → bc<br>C → aAd | enforce rule utility |

**Table 1** Operation of the two grammar constraints, digram uniqueness and rule utility

**2.2 Rule Utility**

Until now, the right-hand sides of rules in the grammar have been only two symbols long. Longer rules are formed by the effect of the rule utility constraint, which ensures that every rule is used more than once. In Table 1, symbol 10 demonstrates this idea. When *d* is appended to rule *S*, the new digram *Bd* causes a new rule, *C*, to be formed. However, forming this rule leaves only one appearance of rule *B*, violating the second constraint. For this reason, *B* is removed from the grammar, and its right-hand side is substituted in the one place where it occurs. Removing *B* means that rule *C* now contains three symbols. This is the mechanism for forming long rules: form a short rule temporarily, and if subsequent symbols continue the match, allow a new rule to supersede the shorter one and delete the latter. Although this creation and deletion of rules appears inefficient at first glance, it can be performed efficiently with the appropriate data structures. More importantly, it keeps track of long matches within the grammar, obviating the need for external data structures. This simplifies the algorithm considerably, permitting a concise proof of its linear time complexity (see Section 5).

## 3. Structures Inferred From Realistic Sequences

Having described the mechanism by which SEQUITUR builds a grammar, and before embarking on efficiency issues, it is instructive to consider the structures that this simple technique can infer from realistic sequences. In each case we applied SEQUITUR to a large sample and then excerpted part of the structure for illustrative purposes.

Figures 2a, 2b and 2c show parts of three hierarchies inferred from the text of the Bible in English, French, and German. The hierarchies are formed without any knowledge of the preferred structure of words and phrases, but nevertheless capture many meaningful regularities. In Figure 2a, the word *beginning* is split into *begin* and *ning*—a root word and a suffix. Many words and word groups appear as distinct parts in the hierarchy (spaces have been made visible by replacing them with bullets). The same algorithm produces the French version in Figure 2b, where *commencement* is split in an analogous way to *beginning*—into the root *commence* and the suffix *ment*. Again, words such as *Au*, *Dieu* and *cieux* are distinct units in the hierarchy. The German version in Figure 2c correctly identifies all words in the sentence, as well as the phrase *die Himmel und die*. In fact, the hierarchy for *the heaven and the* in Figure 2a bears some similarity to the German equivalent.

The London/Oslo-Bergen corpus (Johansson *et al.*, 1978) contains 1.2 million words tagged with word classes. For example, the sentence *Most Labour sentiment would still favour the abolition of the House of Lords* is tagged with the classes *determiner noun noun auxiliary adverb verb article noun preposition article noun preposition noun*. The hierarchy that SEQUITUR infers from the word classes corresponds to a possible parse of each sentence in the corpus, because it is a tree expressed in terms of parts of speech. Figure 2d shows part of the inferred hierarchy, where the tags have been replaced by the actual words from the text. SEQUITUR identifies the middle part of the sentence, *sentiment would still favour the abolition* as a large block, and this part could stand on its own as a grammatical sentence. The adjectival phrase *of the House of Lords* also appears as a distinct unit, as does *Most Labour*, an adjectival phrase that precedes the subject.

Figure 2e shows two Bach chorales from which SEQUITUR has detected both internal repetitions—the light gray boxes show that the two halves of the first chorale are almost identical—and repetitions between chorales, as denoted by the gray box in the second half of the second chorale. The section in the white box occurs in all four halves. Also, by detecting repeated motifs between many chorales, SEQUITUR identifies the imperfect and perfect cadences at the end of the first and second halves, respectively. In general, SEQUITUR is capable of making plausible inferences of lexical structure in sequences, so that the hierarchies it produces aid comprehension of the sequence.

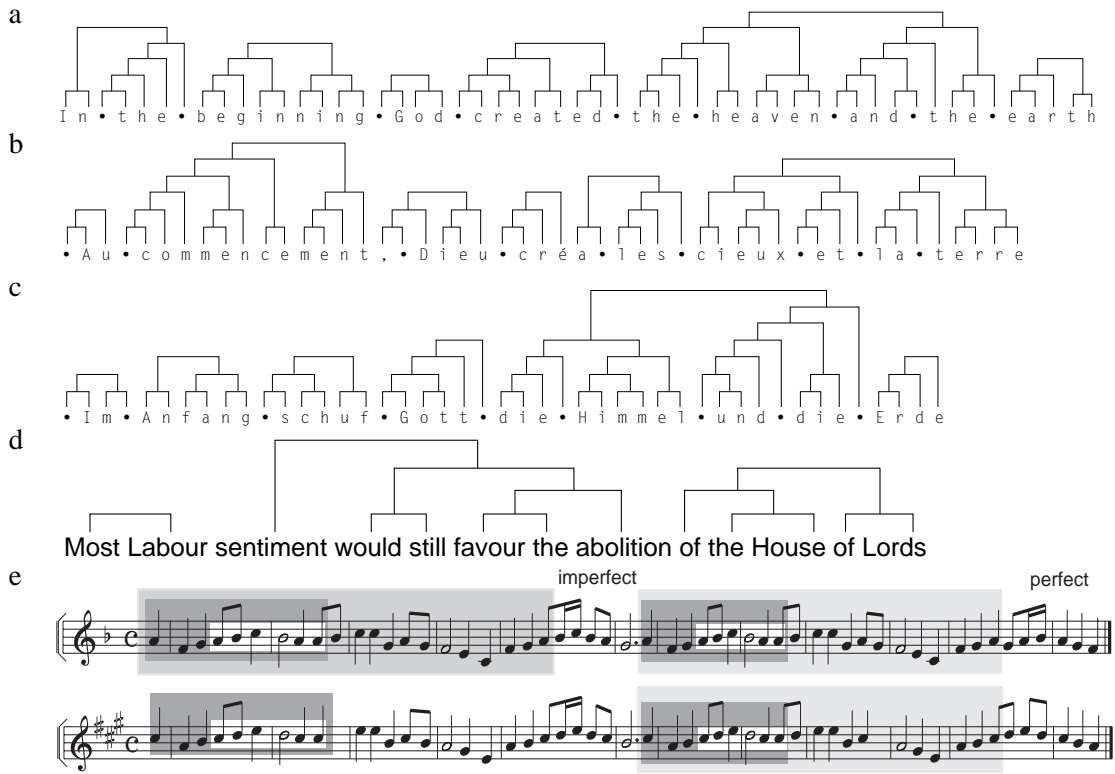

**Figure 2** Hierarchies for various sequences: Genesis 1:1 in (a) English, (b) French, and (c) German; (d) grammatical parse inferred from a sequence of word classes; and (e) repetitions within and between two chorales harmonized by J.S. Bach.

## 4. Implementation Issues

The SEQUITUR algorithm operates by enforcing the digram uniqueness and rule utility constraints. It is essential that any violation of these constraints be detected efficiently, and in this section we will describe the mechanisms that fulfill this requirement.

The choice of an appropriate data structure depends on the kind of operations that need to be performed to modify the grammar. For SEQUITUR these are:
- appending a symbol to rule S;
- using an existing rule;
- creating a new rule; and
- deleting a rule.

Appending a symbol involves lengthening rule *S*. Using an existing rule involves substituting a non-terminal symbol for two symbols, thereby shortening the rules containing the digrams. Creating a new rule involves creating a new non-terminal symbol for the left-hand side, and inserting two new symbols as the right-hand side. After creating the rule, substitutions are made as for an existing rule by replacing the two digrams with the new non-terminal symbol. Deleting a rule involves moving its contents to replace a non-terminal symbol, which lengthens the rule containing the non-terminal symbol; the left-hand side of the rule must then be deleted.

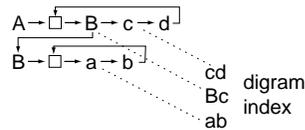

**Figure 3** Data structures for rules and digram index

To ensure that rules can be lengthened and shortened efficiently, SEQUITUR represents a rule using a doubly-linked list whose start and end are connected to a single guard node, shown for two rules *A* and *B* in Figure 3. The guard node also serves as an attachment point for the left-hand side of the rule, because it remains constant even when the rule contents change. Each non-terminal symbol also points to the rule it heads, shown in Figure 3 by the pointer from the non-terminal symbol *B* in rule *A* to the head of rule *B*. With these pointers, no arrays are necessary for accessing rules or symbols, because operations only affect adjacent symbols or rules headed by a non-terminal.

The rule utility constraint demands that a rule be deleted if it is referred to only once. Each rule has an associated reference count, which is incremented when a non-terminal symbol that references the rule is created, and decremented when the non-terminal symbol is deleted. When the reference count falls to one, the rule is deleted.

The digram uniqueness constraint is more difficult to enforce. When a new digram appears, SEQUITUR must search the grammar for any other occurrence of it. One simple solution would be to scan the entire grammar each time looking for a match, but this is inefficient and leads to a quadratic-time algorithm. A better solution requires an index that is efficient to search.

The data structure for storing the digram index must permit fast access and efficient addition and deletion of entries. A hash table provides constant-time access, and adding and deleting entries requires little extra work. Because no digram appears more than once, the table need only contain a pointer to the first symbol of the single matching digram in the grammar data structure, as shown in Figure 3. The hash table consists of a simple array of pointers, and collisions are handled by *open addressing* to avoid the allocation of memory that chaining requires (Knuth, 1968).

Every time a new digram appears in the grammar, SEQUITUR adds it to the index. A new digram appears as a result of two pointer assignments linking two symbols together in the doubly-linked list (one forward pointer and one back pointer). Thus updating the index can be incorporated into the low-level pointer assignments. A digram also disappears from the grammar when a pointer assignment is made—the pointer value that is overwritten by the assignment represents a digram that no longer exists.

To demonstrate the mechanism for updating the hash table when a new rule is created, Table 2 shows the example in Figure 1a, with the addition of the contents of the digram index. When the second *c* is appended to rule *S*, the digram table shows that *bc* already exists in the grammar, so the rule $A \to bc$ is created. Creating the link between *b* and *c* in the right-hand side for rule *A* updates the entry in the index for *bc* to point to its new location—the hash table now contains a pointer to the symbol *b* at the start of rule *A*. Next, the first *bc* is removed. This breaks the link between the *b* in the digram and the preceding symbol *a*, so *ab* is removed from the index. It also breaks the link between *c* and the following *d*, so *cd* is removed from the index. Next, *A* replaces *bc*, creating links between *a* and *A*, as well as between *A* and *d*, adding these digrams to the index. This process continues, resulting in a correct index of digram pointers, but costing just one indexing operation per two pointer operations.

| Action | grammar | change in digrams | digram index |
|---|---|---|---|
| observe symbol 'c' | S → abcdbc | | {ab, bc, cd, db} |
| make new rule A | S → abcdbc<br>A → bc | bc updated | {ab, bc, cd, db} |
| substitute A for bc | S → aAdbc<br>A → bc | ab, cd removed,<br>aA, Ad added | {bc, db, aA, Ad} |
| substitute A for bc | S → aAdA<br>A → bc | db removed,<br>dA added | {bc, dA, aA, Ad} |

**Table 2**  Updating the digram index as links are made and broken

Next, SEQUITUR requires an efficient strategy for checking the digram index. Rechecking the entire grammar whenever a symbol is added is infeasible, and inefficient if large portions of the grammar are unchanged since the last check. In fact, the only parts that need checking are those where links have been made or broken. That is, when any of the actions that affect the maintenance of the digram table are performed, the newly created digrams should be checked in the index. Of course, every time a link is created, the digram is *entered* into the index, and this is the very time to check for a duplicate. If an entry is found to be already present while attempting to add a new digram to the index, a duplicate digram has been detected and the appropriate actions should be performed. Therefore, only one hash table lookup is required for both accessing and updating the digram index.

## 5. Computational Complexity

In this section, we show that the SEQUITUR algorithm is linear in space and time. The complexity proof is an amortized one—it does not put a bound on the time required to process one symbol, but rather bounds the time taken for the whole sequence. The processing time for one symbol can in fact be as large as $O(\sqrt{n})$, where *n* is the number of input symbols so far. However, the pathological sequence that produces this worst case requires that the preceding $O(\sqrt{n})$ symbols involve no formation or matching of rules.

The basic idea of the proof is that the two constraints both have the effect of reducing the number of symbols in the grammar, so the work done satisfying the constraints is bounded by the compression achieved on the sequence. The savings cannot exceed the original size of the input sequence, so the algorithm is linear in the number of input symbols.

| | | action |
|---|---|---|
| 1 | As each new input symbol is observed, append it to rule S. | 1 |
| 2<br>3<br>4<br>5<br>6<br>7<br>8<br>9 | Each time a link is made between two symbols<br>  if the new digram is repeated elsewhere and the repetitions do not overlap,<br>    if the other occurrence is a complete rule,<br>      replace the new digram with the non-terminal symbol that heads the rule,<br>    otherwise,<br>      form a new rule and replace both digrams with the new non-terminal symbol<br>  otherwise,<br>    insert the digram into the index | 2<br><br><br>3<br><br>4 |
| 10<br>11<br>12 | Each time a digram is replaced by a non-terminal symbol<br>  if either symbol is a non-terminal symbol that only occurs once elsewhere,<br>    remove the rule, substituting its contents in place of the other non-terminal symbol | <br><br>5 |

**Table 3**  The SEQUITUR algorithm

Table 3 gives pseudo-code for the SEQUITUR algorithm. Line 1 deals with new observations in the sequence, lines 2 through 9 enforce the digram utility constraint, and lines 10 through 12 enforce rule utility. The on-line appendix contains an implementation of SEQUITUR in Java, which requires about 400 lines for the algorithm.

The numbers at the right of Table 3 identify the main sections of the algorithm, and the proof will demonstrate bounds on the number of times that each of them executes. Action 1 appends symbols to rules $S$ and is performed exactly $n$ times, once for every symbol in the input. Link creation triggers action 2. Action 3 uses an existing rule, action 4 forms a new rule, and action 5 removes a rule.

Table 4 shows examples of actions 3, 4, and 5, and the associated savings in grammar size. The savings are calculated by counting the number of symbols in the grammar before and after the action. The non-terminal symbols that head rules are not counted, because they can be recreated based on the order in which the rules occur. Actions 3 and 5 are the only actions that reduce the number of symbols. There are no actions that increase the size of the grammar, so the difference between the size of the input and the size of the grammar must equal the number of times that both these actions have been taken.

Now that we have set the stage, we can proceed with the proof. More formally, let

$n$ be the size of the input string,
$o$ be the size of the final grammar,
$r$ be the number of rules in the final grammar,
$a_1$ be the number of times new symbol is seen (action 1),
$a_2$ be the number of times a new digram is seen (action 2),
$a_3$ be the number of times an existing rule is used (action 3),
$a_4$ be the number of times a new rule is formed (action 4), and
$a_5$ be the number of times a rule is removed (action 5).

According to the reasoning above, the reduction in the size of the grammar is the number of times actions 3 and 5 are executed. That is,

$$n - o = a_3 + a_5. \qquad (1)$$

Next, the number of times a new rule is created (action 4) must be bounded. The two actions that affect the number of rules are 4, which creates rules, and 5, which deletes them. The number of rules in the final grammar must be the difference between the frequencies of these actions:

$$r = a_4 - a_5.$$

In this equation, $r$ is known and $a_5$ is bounded by equation (1), but $a_4$ is unknown. Noting that $a_1$, the number of times a new symbol is seen, is equal to $n$, the total work is

$$a_1 + a_2 + a_3 + a_4 + a_5 = n + a_2 + (n - o) + (r + a_5).$$

| | action | before | after | saving |
|---|---|---|---|---|
| Matching existing rule | 3 | ...ab...<br>A → ab | ...A...<br>A → ab | 1 |
| Creating new rule | 4 | ...ab...ab... | ...A...A...<br>A → ab | 0 |
| Deleting a rule | 5 | ...A...<br>A → ab | ...ab... | 1 |

**Table 4**  Reduction in grammar size for the three grammar operations

To bound this expression, note that the number of rules must be less than the number of symbols in the final grammar, because each rule contains at least two symbols, so that

$r < o$.

Also, from expression (1) above, we have

$a_5 = n - o - a_3 < n$.

Consequently,

$a_1 + a_2 + a_3 + a_4 + a_5 = 2n + (r - o) + a_5 + a_2 < 3n + a_2$.

The final operation to bound is action 2, which checks for duplicate digrams. Searching the grammar is done by hash table lookup. Assuming an occupancy less than, say, 80% gives an average lookup time bounded by a constant (Knuth, 1967). This occupancy can be assured if the size of the sequence is known in advance, or by enlarging the table and recreating the entries whenever occupancy exceeds 80%. The number of entries in the table is just the number of digrams in the grammar, which is the number of symbols in the grammar minus the number of rules in the grammar, because symbols at the end of a rule do not form the left hand side of any digram. Thus the size of the hash table is less than the size of the grammar, which is bounded by the size of the input. This means that the memory requirements of the algorithm are linear. In practice, the linear growth of memory poses a problem. One strategy that we are currently investigating is to break the input into small segments, form grammars from each of them, and merge the resulting grammar.

As for the number of times that action 2 is carried out, a digram is only checked when a new link is created. Links are only created by actions 1, 3, 4 and 5, which have already been shown to be bounded by $3n$, so the time required for action 2 is also O($n$).

Thus we have shown that the algorithm is linear in space and time. However, this claim must be qualified: it is based on a register model of computation rather than a bitwise one. We have assumed that the average lookup time for the hash table of digrams is bounded by a constant. However, as the length of the input increases, the number of rules increases without bound, and for unstructured (e.g., random) input, the digram table will grow without bound. Thus the time required to execute the hash function and perform addressing will not be constant, but will increase logarithmically with the input size. Our proof ignores this effect: it assumes that hash function operations are register-based and therefore constant time. In practice, with a 32-bit architecture, the linearity proof remains valid for sequences of up to around $10^9$ symbols, and for a 64-bit architecture up to $10^{19}$ symbols.

## 6. Exploring the Extremes

Having described SEQUITUR algorithmically, we now characterize its behavior in a variety of domains. This section explores how large or small a grammar can be for a given sequence length, as well as determining the minimum and maximum amount of work the algorithm can carry out and the amount of work required to process one symbol. Figure 4 summarizes these extreme cases, giving part of an example sequence and the grammar that results. Bounds are given in terms of $n$, the number of symbols in the input.

The deepest hierarchy that can be formed has depth O($\sqrt{n}$), and an example of a sequence that creates such a hierarchy is shown in Figure 4a. In order for the hierarchy to deepen at every rule, each rule must contain a non-terminal symbol. Furthermore, no rule need be longer than two symbols. Therefore, to produce a deep hierarchy from a short string, each rule should be one terminal symbol longer than the one on which it builds. In order to create these rules, the string represented must appear in two different contexts; otherwise the rule will be incorporated into a longer rule. One context is the deepest hierarchy, in which it must participate. The other context should not be in any hierarchy, to reduce the size of the input string, so it should appear in rule *S*. Note that every rule in Figure 4a appears both in the hierarchy and in rule *S*. At each repetition of the sequence, one terminal symbol is appended, producing a new level in the hierarchy. There is no point in including a repetition of length one, so the $m^{th}$ repetition has length $m + 1$. This repetition gives rise to the $m^{th}$ rule (counting rule *S*). The total length of the sequence for a hierarchy of depth *m* is therefore

$$n = 2 + 3 + 4 + ... + (m + 1) = O(m^2)$$

and the deepest hierarchy has depth $m = O(\sqrt{n})$.

At the other end of the spectrum, the grammar with the shallowest hierarchy, shown in Figure 4b, has no rules apart from rule *S*. This grammar is also the largest possible one for a sequence of a given length, precisely because no rules can be formed from it. The sequence that gives rise to it is one in which no digram ever recurs. Of course, in a sequence with an alphabet of size $|\Sigma|$, there are only $O(|\Sigma|^2)$ different digrams, which bounds the length of such a sequence. This kind of sequence produces the worst case compression: there are no repetitions, and therefore SEQUITUR detects no structure.

|   |   | bound | example sequence | example grammar |
|---|---|---|---|---|
| a | deepest hierarchy | $O(\sqrt{n})$ | ababcabcdabcdeabcdef | S → ABCDDf |
|   |   |   |   | A → ab |
|   |   |   |   | B → Ac |
|   |   |   |   | C → Bd |
|   |   |   |   | D → Ce |
| b | largest grammar; shallowest hierarchy | n | aabacadae...bbcbdbe... | S → aabacadae... |
| c | smallest grammar | $O(\log n)$ | aaaaaaaaaaaaaaa... | S → DD |
|   |   |   |   | A → aa |
|   |   |   |   | B → AA |
|   |   |   |   | C → BB |
|   |   |   |   | D → CC |
| d | largest number of rules | n/4 | aaaaababacacadad... | S → AABBCCDD |
|   |   |   |   | A → aa |
|   |   |   |   | B → ab |
|   |   |   |   | C → ac |
|   |   |   |   | D → ad |
| e | maximum processing for one symbol | $O(\sqrt{n})$ | yzxyzwxyzvwxy | S → ABwBvwxy |
|   |   |   |   | A → yz |
|   |   |   |   | B → xA |
| f | greatest number of rule creations and deletions | n new rules n deleted rules | abcdeabcdeabcde... | S → AAA... |
|   |   |   |   | A → abcde |

**Figure 4** Some extreme cases for the algorithm

Turning from the largest grammar to the smallest, Figure 4c depicts the grammar formed from the most ordered sequence possible—one consisting entirely of the same symbol. When four contiguous symbols appear, such as *aaaa*, a rule $B \rightarrow aa$ is formed. When another four appear, rule *S* contains *BBBB*, forming a new rule $C \rightarrow BB$. Every time the number of symbols doubles, a new rule is created. The hierarchy is thus $O(\log n)$ deep, and the grammar is $O(\log n)$ in size. This represents the greatest data compression possible, although it is not necessary to have a sequence of only one symbol to achieve this logarithmic lower bound—any recursive structure will do.

To produce the grammar with the greatest number of rules, each rule should only include terminal symbols, because building a hierarchy will reduce the number of rules required to cover a sequence of a given size. Furthermore, no rule should be longer than two symbols or occur more than twice. Therefore each rule requires four symbols for its creation, so the maximum number of rules for a sequence of length *n* is *n*/4, as shown in Figure 4d.

Having discussed the size of grammars, we can now consider the effort involved in maintaining them. We have shown that the upper bound for processing a sequence is linear in the length of the sequence. However, it is still useful to characterize the amount of processing involved for each new symbol. Figure 4e shows a sequence where the repetition is built up as *yz*, then *xyz*, then *wxyz*, and so forth. Just before the second occurrence of *wxyz* appears, no matches have been possible for the *w*, *x*, and *y*. When *z* appears, *yz* matches rule *A*, then *xA* matches rule *B*. Finally, SEQUITUR forms a new rule for *wB*. This cascading effect can be arbitrarily large if the repetitions continue to build up in this right-to-left fashion. The amount of processing required to deal with the last *z* is proportional to the depth of the deepest hierarchy, as the matching cascades up the hierarchy. The maximum time to process one symbol is therefore $O(\sqrt{n})$. The fact that *w*, *x*, and *y* fail to match means that they require little time to process, preserving the overall linear time bound.

Although the bound is linear, sequences certainly differ in the proportion of work to sequence length. The sequence in Figure 4b, where no repetitions exist and no grammar is formed, minimizes this ratio. The sequence in Figure 4f, which consists of multiple repetitions of a multi-symbol sequence, maximizes it. Each time the repetition appears there are several rule deletions and creations as the match lengthens. In fact, every symbol except *a* incurs a rule creation and a subsequent deletion, so there are $O(n)$ creations and deletions. If *m* is the length of the repetition, the proportion of symbols that do not incur this work is $1/m$, which tends toward zero as the repetition length approaches infinity.

## 7. Behavior in Practice

To give an idea of how SEQUITUR behaves on realistic sequences, we turn from artificial cases to a sequence of English text. Figure 5a plots the number of rules in the grammar against the number of input symbols for a 760,000 character English novel, and shows that the increase is approximately linear. Figure 5b shows the approximately linear growth of the total number of symbols in the grammar. The growth of the number of unique words in the text, shown in Figure 5c, is high at the start and drops off toward the end. Zobel, *et al.* (1995) have observed in much larger samples of English text that—surprisingly—new words continue to appear at a fairly constant rate, corresponding not just to neologisms but to names, acronyms, and typographical errors. In this example, the number of rules grows linearly because, once words have been recognized, multi-word phrases are constructed, and the number of such phrases is unbounded.

The nearly linear growth of the number of symbols in the grammar seems disappointing, but is in fact an inevitable consequence of the information content of English. Since symbols at the end of the text convey a similar amount of information as symbols at the beginning, there is a lower bound on the achievable compression rate. For English text, this corresponds to the entropy of English.

SEQUITUR operates very quickly—as shown in Figure 5d, the 760,000 character novel is processed in 16 seconds: a rate of 50,000 symbols per second, or 3 Mb per minute. The figure also illustrates SEQUITUR's linear-time behavior in practice. The sequence in Figure 4b, where no repetitions exist and no rules are formed, should be fast to process, and indeed it is processed at a rate of 150,000 symbols per second—three times faster than the novel. The sequence in Figure 4f, which consists of multiple repetitions of a multi-symbol sequence, slows performance to 14,000 symbols per second—a ten-fold decrease from the fastest sequence. The sequence in Figure 4c, which consists of many repetitions of a single character and forms a concise grammar, comes in at 50,000 symbols per second; about the same as the novel. All measurements were performed on a Silicon Graphics Indigo 2.

SEQUITUR is an effective data compression scheme that outperforms other schemes that achieve compression by factoring out repetition, and approaches the performance of schemes that compress based on probabilistic predictions. SEQUITUR's implementation and evaluation as a compression scheme is described by Nevill-Manning and Witten (1997).

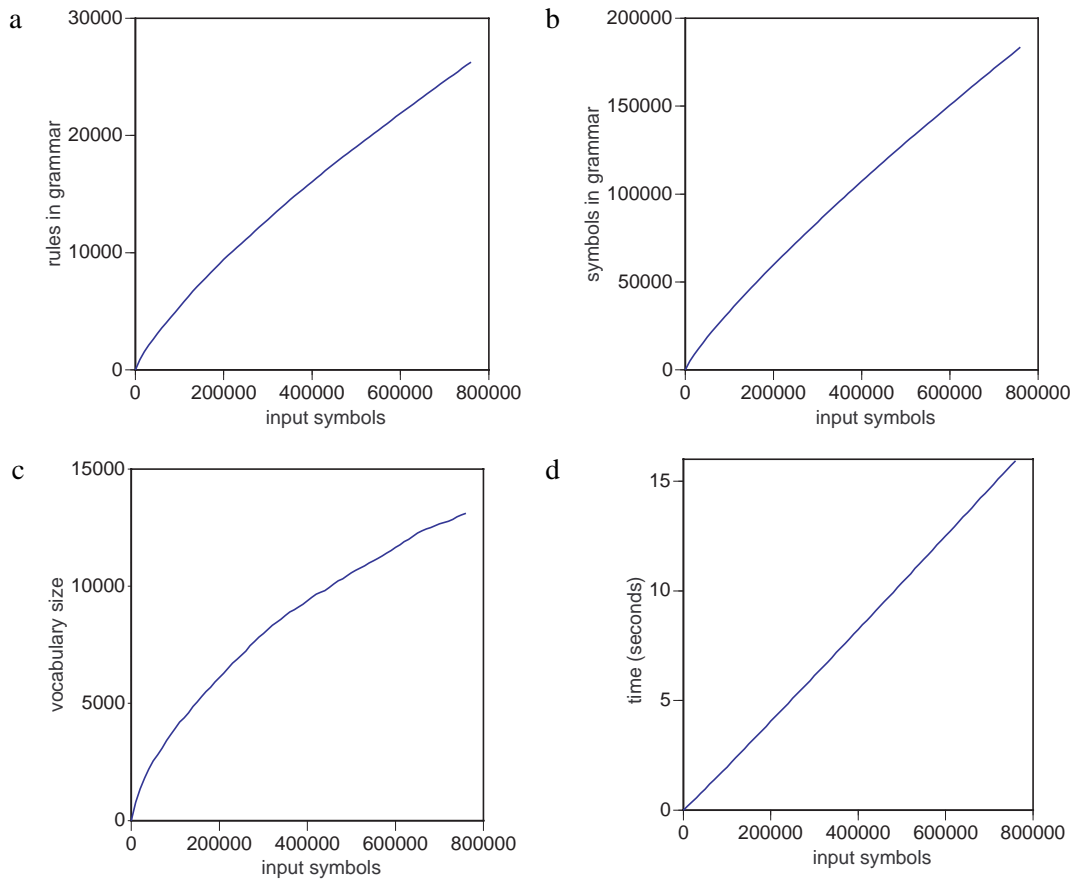

**Figure 5** Growth rates on English text: (a) rules in the grammar; (b) symbols in the grammar; (c) vocabulary size in the input; and (d) time

## 8. Related Work

As mentioned in the introduction, this research resembles work by Wolff (1975). Having described SEQUITUR, it is now possible to contrast it with Wolff's system, MK10, which processes a sequence from left to right, and forms a 'chunk' (equivalent to a SEQUITUR rule) whenever a digram is seen more than 10 times. When this happens, all occurrences of the digram are replaced with a non-terminal symbol, and the system either carries on in the sequence, or restarts from the beginning. In either case, digram frequencies are discarded and the process starts over. The worst case for this algorithm corresponds to the sequence in Figure 4f, where there are long exact repetitions. Each symbol in the repeated segment gives rise to a chunk, and the process starts over. In Figure 4f, the length of the repetition is linear in the length of the sequence, and the number of restarts is the length of the repetition, so the algorithm is quadratic in the length of the sequence. This makes processing of million-symbol sequences impractical.

A number of systems, by Langley (1994), Stolcke and Omohundro (1994), and Cook *et al*. (1976), form new grammar rules from repeated sequences, and also merge rules to generalize grammars. However, they operate in a different domain—as input, they expect a set of sentences drawn from a language, rather than a single long sequence. This allows them to make inferences based on directly comparing corresponding parts of different sequences. Furthermore, the small size of the training data means that efficiency is of lesser concern. The performance of these algorithms is measured by their ability to accept test sentences from the language, and to reject new sentences that are not in the target languages. In SEQUITUR's case, where there is only one sequence available, this metric does not apply.

VanLehn and Ball (1987) also infer grammars from sets of sentences. Their algorithm enforces three constraints on grammars for the purpose of making a version space finite. They are: "(1) no rule has an empty right side, (2) if a rule has just one symbol on its right side, the symbol is a terminal, and (3) every non-terminal appears in a derivation of some string." These constraints are reminiscent of SEQUITUR's—for example, the third constraint is a weaker form of SEQUITUR's rule utility—but serve a different purpose. In SEQUITUR, they are operational; they drive the formation of the grammar. In VanLehn and Ball's work, they make the version space tractable by providing sensible restrictions on the form of the grammar, and the algorithm itself is a search through the space.

## 9. Conclusion

This paper has presented SEQUITUR, an algorithm for identifying hierarchical structure in sequences. Based on the idea of abstracting subsequences that occur more than once into rules and continuing this operation recursively, the algorithm works by maintaining two constraints: every digram in the grammar must be unique, and every rule must be used more than once. SEQUITUR operates incrementally and, subject to a caveat about the register model of computation used, in linear space and time. This efficiency has permitted its application to long sequences—up to 40 Mbyte—in many different domains.

We have not evaluated the prediction accuracy of SEQUITUR in this paper. Evaluating prediction accuracy is a fairly complex business. It is not adequate simply to give a count of correct versus incorrect predictions, because doing this begs the question of the likelihood of different ones occurring. Prediction schemes can assign probabilities to all the predictions they might offer, and should be judged on the discrepancy between their probabilistic predictions and the true upcoming symbols. The whole question of accurate probabilistic prediction of sequences is tantamount to the compression of sequences, a substantial field in its own right (Bell *et al.*, 1990). We have in fact evaluated SEQUITUR's performance in compression and found that it vies with the best compression algorithms, particularly when a large amount of text is available (Nevill-Manning and Witten, 1997). But the point of the present paper is a different one: that SEQUITUR re-represents a sequence in a way that exposes its underlying structure. It is fair to say that no other compression algorithm produces a representation that is in any way perspicuous.

Perhaps the greatest drawback to the SEQUITUR algorithm is its memory usage, which is linear in the size of grammar. Linear memory complexity is ordinarily considered intractable, although in practice SEQUITUR works well on sequences of rather impressive size. There is clearly room for approximate versions of the algorithm that partition the input and re-merge the grammars formed from them, and this could perhaps be applied recursively to create an algorithm with logarithmic memory requirements. We conjecture, however, that while such approximations will no doubt turn out to be very useful in practice, they will inevitably sacrifice the property of digram uniqueness that is an appealing feature of the original algorithm.

## Acknowledgments

We are grateful for many detailed suggestions from Pat Langley and the anonymous referees.

## References

Andreae, J.H. (1977) *Thinking with the teachable machine*. London: Academic Press.

Angluin, D. (1982) Inference of reversible languages, *Journal of the Association for Computing Machinery, 29*, 741–765.

Bell, T.C., Cleary, J.G., and Witten, I.H. (1990) *Text compression*. Englewood Cliffs, NJ: Prentice-Hall.

Berwick, R.C., and Pilato, S. (1987) Learning syntax by automata induction, *Machine Learning, 2*, 9–38.

Cohen, A., Ivry, R.I., and Keele, S.W. (1990) Attention and structure in sequence learning, *Journal of Experimental Psychology, 16*(1), 17–30.

Cook, C.M., Rosenfeld, A., & Aronson, A. (1976). Grammatical inference by hill climbing, *Informational Sciences*, *10*, 59-80.

Cypher, A., editor (1993) *Watch what I do: programming by demonstration*, Cambridge, Massachusetts: MIT Press.

Gaines, B.R. (1976) Behaviour/structure transformations under uncertainty, *International Journal of Man-Machine Studies, 8*, 337–365.

Gold, M. (1967) Language identification in the limit, *Information and Control, 10*, 447–474.

Johansson, S., Leech, G., and Goodluck, H. (1978) "Manual of Information to Accompany the Lancaster-Oslo/Bergen Corpus of British English, for Use with Digital Computers," Oslo: Department of English, University of Oslo.

Knuth, D.E. (1968) *The art of computer programming 1: fundamental algorithms.* Addison-Wesley.

Laird, P. & Saul, R. (1994) Discrete sequence prediction and its applications, *Machine Learning* 15, 43–68.

Langley, P. (1994). Simplicity and representation change in grammar induction. Unpublished manuscript, Robotics Laboratory, Computer Science Department, Stanford University, Stanford, CA.

Nevill-Manning, C.G. & Witten, I.H. Compression and explanation using hierarchical grammars, *Computer Journal*, in press.

Nevill-Manning, C.G. (1996) *Inferring sequential structure*, Ph.D. thesis, Department of Computer Science, University of Waikato, New Zealand.

Nevill-Manning, C.G., Witten, I.H. & Paynter, G.W. (1997) Browsing in digital libraries: a phrase-based approach, *Proc. Second ACM International Conference on Digital Libraries*, 230–236, Philadelphia, PA.

Rabiner, L.R. and Juang, B.H. (1986) An introduction to hidden Markov models, *IEEE ASSP Magazine, 3*(1), 4–16.

Stolcke, A., & Omohundro, S. (1994). Inducing probabilistic grammars by Bayesian model merging. *Proc. Second International Conference on Grammatical Inference and Applications*, 106–118, Alicante, Spain: Springer-Verlag.

VanLehn, K., & Ball, W. (1987). A version space approach to learning context-free grammars. *Machine Learning*, *2*, 39–74.

Wharton, R. M. (1977). Grammar enumeration and inference. *Information and Control*, 33, 253-272.

Wolff, J.G. (1975) An algorithm for the segmentation of an artificial language analogue, *British Journal of Psychology, 66*, 79–90.

Wolff, J.G. (1977) The discovery of segments in natural language, *British Journal of Psychology, 68*, 97–106.

Wolff, J.G. (1980) Language acquisition and the discovery of phrase structure, *Language and Speech, 23*(3), 255–269.

Wolff, J.G. (1982) Language acquisition, data compression and generalization, *Language and Communication, 2*(1), 57–89.